# Affordance-Based Manipulation Planning with Text Goals and Sim-to-Real Generalisation via Real-to-Sim Image Conversion

Solvi Arnold, Rin Karashima, Tadashi Adachi, Takafumi Mochizuki, Kimitoshi Yamazaki

*Abstract*—We present a manipulation planning system based on affordance recognition and action effect prediction. The system reasons through possible futures in visual form, and evaluates candidate plans by agreement of predicted outcomes with text-based goals set at run-time, using a multimodal goal-matching module. Positions of objects named in the goal text are tracked through predictions even when occluded, making it possible to generate action plans even when objects become occluded, or when their initial descriptors cease to identify them in future states. We further expand the system with an image conversion module for translating real-world state images with objects of varied shapes and visual appearances into a consistent visual appearance, to facilitate manipulation planning in a physical robot setup. We evaluate performance of the system's modules in isolation, and demonstrate the integrated system's manipulation planning capabilities on a set of challenging tasks in both simulation and on hardware.

## I. INTRODUCTION

ACTION planning has traditionally often been addressed using symbolic methods, and more recently using LLMs. Such methods can be effective if a symbolic or linguistic description can adequately convey all the relevant features of the task environment. However, this is rarely the case in real-world manipulation scenarios, limiting the planning fidelity we can achieve using symbolic methods. Furthermore, when effect rules are defined manually, planning ability is constrained by the imagination of the rule-designer, resulting in a system than cannot reliably avoid or exploit effects and side-effects that the designer overlooked. Symbolic reasoning certainly has a role in human action planning, but our reasoning is built on top of a more granular understanding of the physical world.

In a robotics context, which actions are available in a given situation intricately depends on the robot's physical characteristics and limitations. Furthermore, for an action to be executable and reliably produce the intended effects, it has to be parametrised in a manner that is translatable to unambiguous motions of the target robot platform. Hence, for a robot to use its action repertoire fully and reliably, it must be able to 1) recognise which actions are available to it in a given scene (i.e. recognise the scene's *affordances*), and 2) accurately anticipate the effects of those actions.

The abovementioned concept of affordances [1] has proven useful for structuring robot behaviour. Assigning regions of robots' (typically continuous) state spaces to qualitatively distinct affordances facilitates the formation of meaningful relations between perception and action spaces. However, the effects of executing affordances so defined are less straightforward to model. We may assume to know our robot in detail, but the same assumption can generally not be made for the environments in which it will operate. Similar observations apply to biological agents. For humans and other animals, nature's approach has been to equip us with the ability to learn the effects of our actions from experience.

Based on the above, [2] introduced a manipulation planning method combining recognition of prespecified affordances with learned prediction of affordance effects, both implemented as neural networks (NNs). The present work builds on [2], addressing some of its limitations. In [2], goals are defined using images, which is inconvenient in practice. We introduce support for text-based goal definition, using a multimodal NN. We relax the assumption that objects are known (i.e. observed during training). Instead, we target varied instances from a set of object *classes* instead. We introduce a *style transfer* approach using generative augmentation to allow the system to generalize over variation in object shape, size, and visual appearance within these classes and enable real-world experiments. We introduce a dynamic camera viewpoint into the prediction process to improve generalization over object positions and orientations. We enrich state representations with object mask channels to keep track of objects named in the goal text, and enable planning for tasks involving full object occlusion.

## II. RELATED WORK

Learning of action effects has been explored in various works. Early examples are found in [3] [4], where transition rules are learned in symbolic form. This avoids the need to manually specify effects in advance, but the granularity of the predictions is similar to that of manually defined rules. In [5], fine-grained prediction for a reachability affordance is learned using a Gaussian Mixture Model. This predictive ability is used to plan reaching motions by means of *internal rehearsal*.

SA is at NICT, Japan. Work performed while at Shinshu University, Japan. (solvi.arnold@nict.go.jp). RK, TA, TM are at EPSON Avasys, Japan. KY is at Tohoku University, Japan. Work performed while at Shinshu University, Japan.

[6] focuses on the formation of compound objects resulting from stacking actions applied to objects of various shapes. The method employs neural prediction of action effects by means of a graph neural network. However, state and effect descriptors are comparatively high-level and task-specific, and state context beyond the focal objects is not considered. More granular prediction for object stacking is found in [7], albeit for simpler objects.

Neural prediction through multi-step action sequences is found in [8]. However, future states are only represented in latent form, necessitating task-specific reinforcement learning. [9] similarly employs a predictive neural architecture operating in latent space, but with focus on using the effects of partial actions.

A new element of our system is the addition of object masks to track specific objects. Prediction over object masks encoding some semantic information was explored in [10], albeit at the single-object level and without prediction of future states' RGBD appearance.

In [11], incremental learning of skill effects is applied to gradually expand a robot's action repertoire. However, states are represented in a manually designed analytical format.

Our work has parallels to [12] in the use of a multimodal NN to assess action quality. However, in [12], such a model is used to assess success probabilities over sets of candidate push actions in order to select the optimal action. In the present work, we use a multimodal NN to assess goal agreement over sets of state predictions. This allows quality assessment over action *sequences* spanning various affordance types.

The combination of affordance-based action generation with language instruction is also explored in [13]. However, the system in [13] expects direct action instruction (the user tells the system what to do), whereas we consider instruction by specifying the desired outcome (the user tells the system what to achieve, but not how to achieve it). Consequently, we relate text to states instead of actions. [14], too, integrates natural language instruction, and solves multi-step tasks with comparatively high-level instruction, but still lets the user specify the solution strategy, rather than stating the goal to achieve.

The use of vision-language models for success detection over visual states and text-based goal specifications were explored in [15]. Success determination was used to provide reward signals for Reinforcement Learning agents. In contrast, we determine success not on the present state but on predicted future states to determine plan quality.

## III. Contributions

1. We propose a manipulation planning system that reasons through sequences of affordances using fine-grained visual state representations, while allowing intuitive goal specification in text format.
2. We integrate a real-to-sim style-transfer approach to enable operation in a real-world setup.
3. We expand the prediction-based planning logic of [2] to enable planning with goals in which key objects become fully occluded and/or stop being identifiable by their initial identifiers, by enriching state representations with object-tracking channels that keep track of objects named in the goal text.
4. We integrate orthogonalisation and virtual camera motion into the system's prediction module to improve generalisation over object layouts.

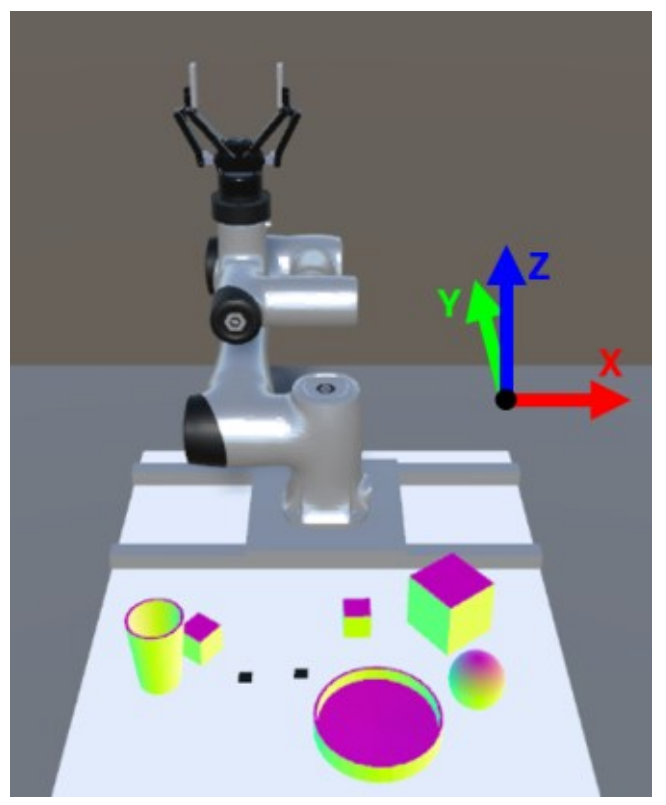

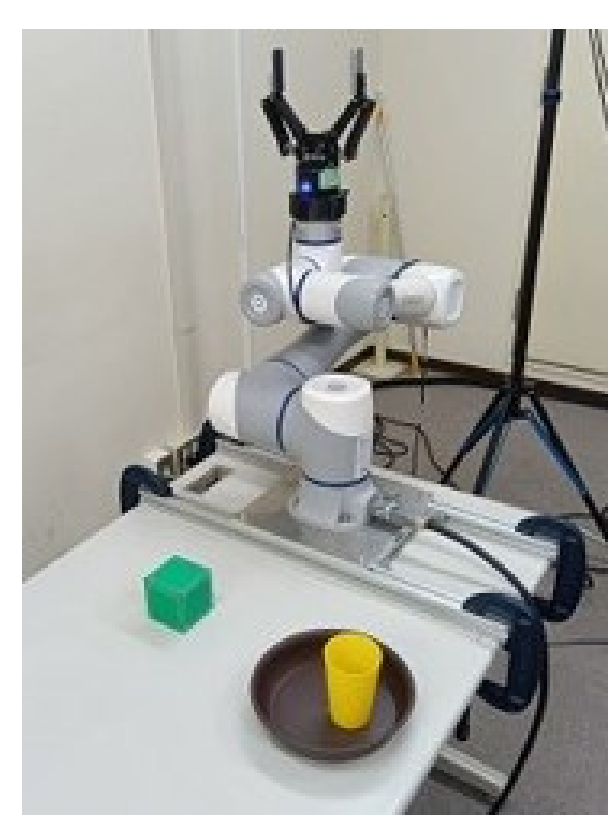

Fig. 1. Left: Simulated task environment (Unity). Right: Real-world task environment.

## IV. Task Environment

We start by explaining the task environment we use to train and evaluate the system. We expand the tabletop manipulation environment of [2] as illustrated in Fig. 1 (left). The scene is built in Unity [16] and features a virtual UR5 robot arm (Universal Robotics), controlled using Unity-ROS infrastructure provided by the Unity Robotics Hub [17], and objects from the following categories: balls, cubes, cups, plates. For data generation, object arrangements are generated randomly, while for evaluation, arrangements are designed manually to create specific planning tasks. For real-world experiments, we closely replicate the simulation environment in a physical setup as shown in Fig. 1 (right).

We define the affordance type repertoire $\mathcal{A}$ as $\{GRASP, PLACE, PUSH, INSERT\}$. Each affordance is parametrised by a pose consisting of 3D coordinates specifying the affordance's position in space, and a single angle specifying a gripper orientation in the XY plane. $GRASP$ affordances can exist on objects that fit within the open gripper in a down-pointing pose, and on rims of upward-facing hollow objects (cups and plates). When the robot is holding an object, $PLACE$ affordances can exist on top of cubes and at freely placed small black markers on the table surface, and $INSERT$ affordances can exist on upward-facing hollow objects. $PUSH$ affordances can exist on the outer surface of the top-down silhouette of an object. With regard to the gripper angle in the affordance parametrisation, we distinguish between objects that present round top-down silhouettes (balls, upright cups, plates), and objects with trapezoid top-down silhouettes (cubes and toppled cups). In the former case the object provides no distinct privileged directions, so we place affordances at 0, 90, 180, and 270° angles. For the latter case, we determine which two of the of the three axes of the object's local coordinate system are least inclined from the work surface, project these axes onto the XY plane, and place affordances in the corresponding directions. This ensures that e.g. cubes will be grasped and pushed at their faces, rather than at precarious oblique angles.

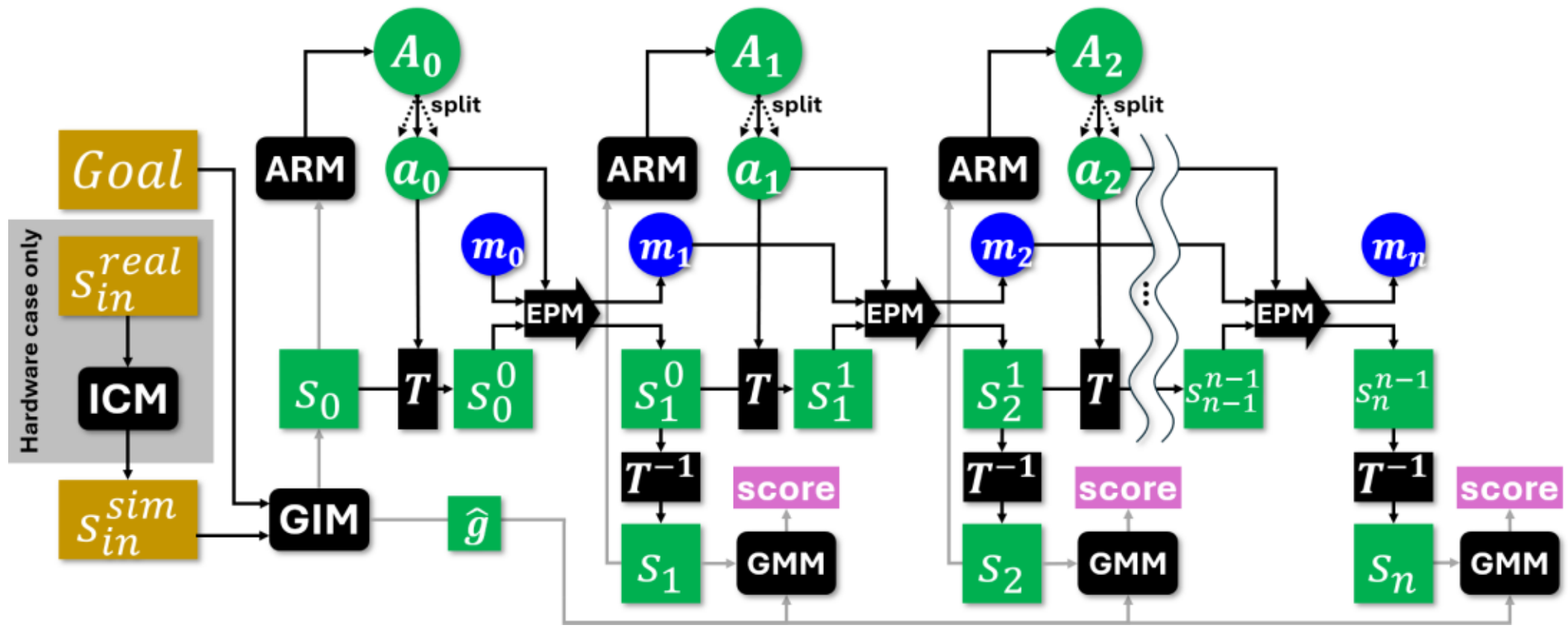

Fig. 2. Global system architecture. ICM: Image Conversion Module. GIM: Goal Interpretation Module. ARM: Affordance Recognition Module. EPM: Effect Prediction Module. GMM: Goal Matching Module. $\boldsymbol{T}$ applies and $\boldsymbol{T}^{-1}$ undoes view transformations. Plan search process branches at the selection of $\boldsymbol{a}_i$ from the list of available affordances $\boldsymbol{A}_i$ (marked "split").

Whether an affordance actually exists at a given pose is determined programmatically during data generation (see below), and is up for determination by the affordance recognition NN during manipulation planning.

### A. Data Generation

To generate data for training the system's various modules, we generate random object layouts. Objects are generated programmatically with randomised parametrisation for size and shape features, within predefined ranges per object category. For balls and cubes, we randomise a single size parameter. For cups and plates, we randomise top radius, bottom radius, and height. All objects are generated as parametrisations of a frustum with optional curved sides and an optional top cavity. This allows generation of a much wider variety of shapes, but we constrain the shape repertoire to the aforementioned set of recognisable objects, to facilitate text description of objects in other parts of the system.

We procedurally detect the affordances that exist in each randomly generated scene. During data generation we have full access to object positions, so we can determine the set of poses at which affordances may exist. We then apply the following checks to determine which of the potential affordances in this set do in fact exist. For $GRASP$ affordances, we place a simplified open gripper at the candidate pose, and check for collisions with surrounding objects. For $PLACE$, we place a gripper holding the currently held object at the target pose, and check for collisions. For $INSERT$, we drop a small ball above the centre of the insertion target, and check whether it remains in contact with the target once its position has stabilised. If the objects remain in contact, we run the same check as for $PLACE$. For push, we place a half-open gripper at the starting pose for the pushing motion, and check for collisions. We then place the gripper and a copy of the target object at the end position of the push motion, and check for collisions again. In each case, if the relevant poses are collision free, the affordances are deemed to exist.

For each randomly generated object layout, we execute a sequence of up to six affordances, detecting the set of affordances in each layout programmatically and randomly selecting affordances for execution. For each sequence, we store the object configuration at each step, the set of affordances detected in each layout, and the affordance actually executed at each step. The object layout here consists of the following: a top-down RGBD image of the scene, top-down depth images of each object rendered separately, and a list of centre coordinates for each object in the scene.

We render all objects using a custom shader that maps objects' surface normal vectors to RGB values. This lets us encode rich shape information into the visual appearance. For purpose of generalisation over object rotation and translation in the effect prediction NN, we want to be able to reposition the camera during neural processing of the state data. Specifically, we want to enable translation of the camera parallel to the camera plane, and rotation around its principal axis. Such rotations correspond straightforwardly to image rotations, but in the usual perspective projection, camera translations introduce parallax changes that are not replicated by simple image translations. To avoid changes in object appearance due to camera translation, we capture all state images in orthogonal projection, so that image translations correspond straightforwardly to camera translations.

## V. System Architecture

The system is comprised of the following modules: the Goal Interpretation Module (GIM), the Affordance Recognition Module (ARM), the Effect Prediction Module (EPM), the Goal-Matching Module (GMM), and (for hardware experiments) the Image Conversion Module (ICM). Fig. 2 details the information flow between the modules during planning. Below we explain the modules individually. Additional information about the architectures and training procedure hyperparameter settings of the NNs used in the various modules are given in the Appendix. All state images are processed in a resolution of 128×128.

*A. Goal-Interpretation Module (GIM)*

The GIM takes as input the initial state representation and the goal text provided by the user. It enriches the state representation with object mask channels that identify the objects named in the goal text, and converts the goal text into a standardised format that relates the named objects to these mask channels.

The text-based goal definition format currently covers objects' absolute and relative positions (including stackings), and which object (if any) is being held by the robot. Table 1 details the goal text vocabulary. An example of a possible goal text is *"[the small ball] is on the right of [the large cube]. [the large cube] is on the bottom left."* The *named objects* in this goal text are *"the small ball"* and *"the large cube"*. Goal text $g$ is standardised by replacing named objects with numbers: *"#1 is on the right of #2. #2 is on the bottom left."*. We denote the standardised goal text as $\hat{g}$ in Fig. 2.

We obtain mask images for the named objects using a semi-automated process. The user is presented with a view of the initial state, and asked to mark each named object with one or more points. We then employ SAM [18] to obtain pixel-level binary-valued object masks for the marked objects. SAM then produces three candidate masks per object. We present the obtained candidate masks to the user and let them select the best candidate mask. In practice, this is usually the top candidate produced by SAM. To capture occlusion relations in the input state, we attenuate the intensity of the mask image from 1 to 0.5 for pixels where the object is present but occluded.

Object masks for objects renamed as #1, #2, #3 in the standardised goal text are concatenated in order to the RGBD state representation, creating an unambiguous correspondence between standardised object names and mask channels. The number of mask channels is fixed at three. When there are fewer than three named objects, the remaining mask channels remain empty (i.e. all zero). We denote the resulting state image format as RGBDMMM, with RGB denoting the colour channels, D the depth channel, and each M indicating a mask channel.

At present we rely on user input to relate object descriptions to objects, but this could likely be automated for most cases using well-established object recognition methods.

*B. Affordance Recognition Module (ARM)*

This module is responsible for discovering the affordances that exist in a given scene. Given an RGBD image of a world state, the module provides a list of affordances existing in that state. The module is substantially the same as in [2]. The implementation is a modified ScaledYOLOv4 [19] network, reworked to recognise parametrised affordances instead of objects. The input image format is extended from YOLO's RGB to RGBD. YOLO's output originally consists of 2D bounding box coordinates along with the bounding boxes' width and height. The modified network replaces this output format with affordance poses as described in Section 4 (i.e. 3D coordinates and a gripper angle).

The module is trained on tuples of world states and affordance lists from the dataset generated in section 4A, using the ScaledYOLOv4 training procedure with minor hyperparameter modifications. Training employs various types of augmentation (see Appendix). We newly introduce an *artifact* augmentation. An artifact observed in some of our future state predictions are low-intensity leftovers of moved objects from previous states. To robustify affordance recognition against such artifacts, we blend in preceding states at random low intensities as an augmentation.

*C. Effect Prediction Module (EPM)*

This module is responsible for predicting the effects of the affordances found by the ARM. Given a world state $s_0$ in RGBDMMM format and a sequence of parametrised affordances $a_{0:n-1}$, the prediction module predicts the state sequence $s_{1:n}$ that would result from executing sequence $a_{0:n-1}$ starting from state $s_0$. During planning, mask channels of the input state are populated with the masks provided by the GIM. During training, mask images of random subsets of the scene objects are used. Prediction includes the mask channels, allowing us to keep track of the named objects explicitly. Mask output captures occlusions in the same format as used in the input (i.e. 1 for present-and-visible, 0.5 for present-but-occluded, 0 for absent).

As noted above, we do not assume a fixed set of objects, but a set of parametrised object *classes*. This dramatically increases the shape variation of the objects encountered. Additionally, object placement is far less constrained than in [2]. We found that training effect prediction as in [2] quickly becomes infeasible in this case. To enable prediction of sufficient quality for planning under the above conditions, we exploit the translational and rotational invariance of affordance effects. The local context in which an affordance is executed can affect its outcome, but absolute position and rotation in the XY plane do not. This allows us to predict the local effects of an affordance execution relative to a coordinate system positioned at the affordance's position and orientation. However, affordances also have effects that are best captured in the robot's fixed coordinate system. Specifically, grasping an object produces a state wherein the robot holds the object in a default holding pose. In this state, it is essential to consider the visual characteristics of the held object to determine the effect of subsequent $PLACE$ and $INSERT$ affordances.

To accommodate prediction using dual viewpoints, we let the prediction module process states as follows. The state image is split into a "robot view" (top part of the state image) and a "table view" (bottom part of the state image). Using the pose information from the affordance input, the table view is shifted and rotated such that the affordance is centred at the image centre and oriented at an angle of 0. Because state images are represented in orthogonal projection, this transformation is equivalent to aligning the camera to the affordance's XY coordinates. The robot view is left as-is. To avoid loss of image content when recentring the table view, we pad the image to 200×200. The two resulting RGBDMMM images are concatenated along the channel dimension and processed as a 14-channel image. We let $s_i^j$ denote the $i^{th}$ state in a sequence with its table view represented in the coordinate system of the

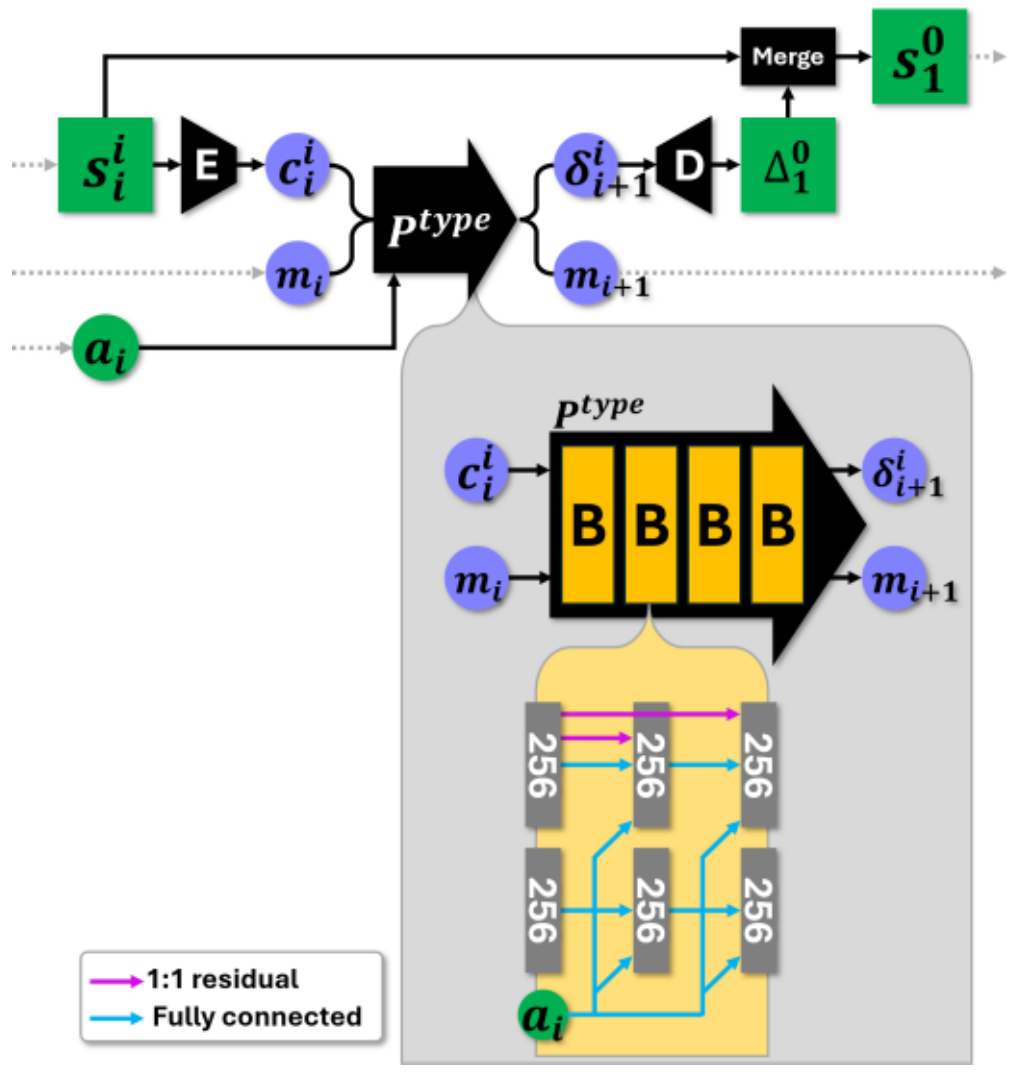

Fig. 3. Architecture of the Effect Prediction Module (EPM).

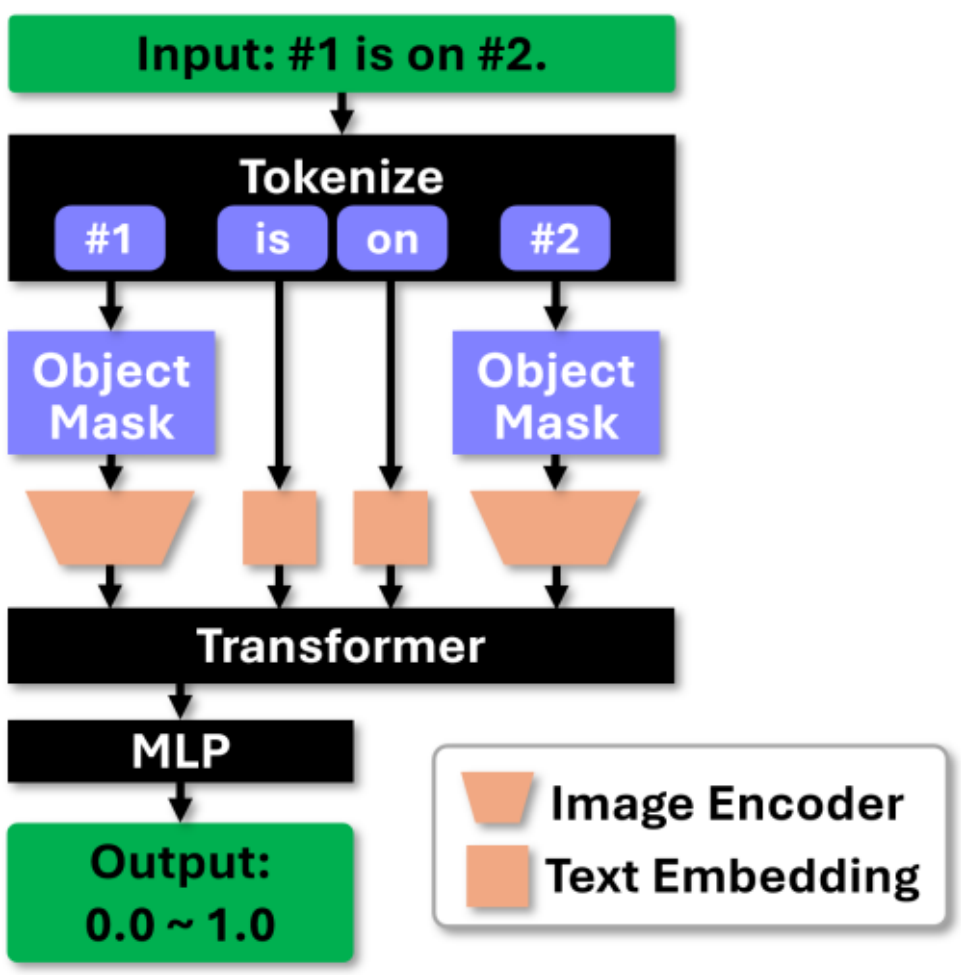

Fig. 4. Architecture of the Goal-Matching Module (GMM) with an example of sentence tokenisation and processing.

$j^{th}$ affordance.

The NN at the core of the EPM expands on [2]. The network consists of encoder $E$, decoder $D$, and predictors $P^{type}$, $type \in \mathcal{A}$. Fig. 3 illustrates the input and output to each subnetwork, and the internal structure of $P^{type}$. Identical architecture is used for each affordance type. Letting $E$ encode $s_i^j$ yields its latent representation $c_i^j$. Given $c_i^i$ and affordance $a_i \in a_{0:n-1}$ of type $type$, $P^{type}$ computes a latent state differential $\delta_{i+1}^i$ representing the predicted differential from state $s_i^i$ to $s_{i+1}^i$, and a memory trace $m_i$. $D$ decodes $\delta_{i+1}^i$ to a (14+2)-channel image $\Delta_{i+1}^i$, consisting of content for the 14 original channels and 2 channels (one per view) providing an "update mask" that determines how this content is combined with the content at the previous step. Given previous channel content $h_i$, new content $h_{i+1}$, and update mask $u$, the resulting channel content is given by $(1-u)h_i + uh_{i+q}$. The resulting 14-channel state image $s_{i+1}^i$ is the predicted state, with its table view represented in the coordinate system of the $i^{th}$ affordance.

As $\delta_{i+1}^i$ is a latent representation, we cannot analytically apply shifts or rotations to the representation of the table view contained therein. For $s_{i+1}^i$, however, we can. Hence, we can undo all shifts and rotations applied so far in order to construct single-view (7-channel) scene representation $s_{i+1}$. During planning, we construct $s_{i+1}$ and pass it to the recognition network in order to obtain the list of affordances available therein, and to the goal matching module to quantify agreement with the planning goal. The single-view representation is also convenient for visualising predicted states for the user. If there is a subsequent affordance $a_{i+1}$, we use the positional and rotational difference between $a_i$ and $a_{i+1}$, to shift and rotate the table view of $s_{i+1}^i$ into the pose of $a_{i+1}$ to obtain $s_{i+1}^{i+1}$, and continue the prediction process. In Fig. 2, $T$ and $T^{-1}$ indicate the application and undoing of viewpoint transformations described above. We use the PIX library in the JAX ecosystem to implement image rotations as part of the EPM's computational graph, so as to allow training gradients to pass through the image rotations. This makes it possible to train the EPM over sequential affordance executions despite the intervening image transformations.

Table 1. Goal sentence patterns.

| Pattern [count] | Meaning |
|---|---|
| {#m} is grasped. [47685] | The robot is holding object **#m**. |
| {#m} is on {#n}. [190168] | Object **#m** is sitting on top of **#n**. |
| {#m} is on the {row} {col}. [590351] | Object **#m** is located in the indicated region of the work desk. **row** ∈ {center, right, left} **col** ∈ {middle, top, bottom} |
| {#m} is {direction} {#n}. [1858014] | Position of **#m** relative to **#n**. **direction** ∈ {above, above left of, left of, below left of, below, below right of, right of, above right of} |

If there is a subsequent affordance in the sequence being processed, we carry the memory trace $m_i$ through as input for the next prediction step. The memory trace allows for retention of information about delayed effects (i.e. effects not yet apparent in $s_{i+1}^i$). The initial memory trace $m_i$ is initialised to all-zero. We note that the experiments in the present work are not designed to require consideration of such effects.

The effect prediction module is trained on the sequences of states and executed affordances from the dataset described in section 4A.

*D. Goal-Matching Module (GMM)*

This newly introduced module is responsible for quantifying the agreement of predicted states with the goal text. Core of the module is a multi-modal NN [20] [15]. Its global architecture is shown in Fig. 4. Input mixes image and text features, and output is a real value in the [0,1] range quantifying text-state agreement. The encoders and transformer derive from CLIP [21]. The image encoder and transformer are scaled down and trained from scratch, while our text encoder is a fine-tuning of CLIP's text embedder. The MLP module is implemented and trained from scratch.

Input to the matching module is a $\langle masks, goal \rangle$ tuple,

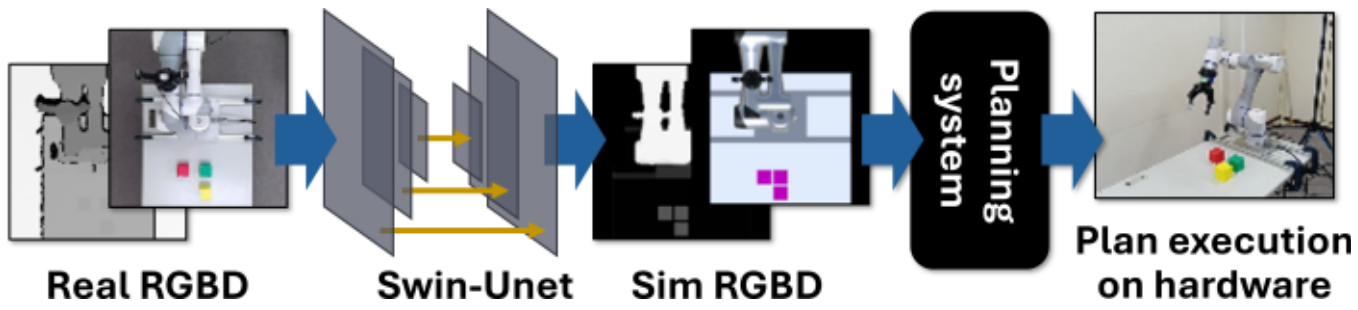

Fig. 5. Concept of domain adaptation via image conversion.

where $masks$ consists of the mask channels of the state image, and $goal$ is a list of sentences. Given a goal sentence such as "*#1 is on #2.*", each named object $\#i$ is replaced with the mask image $v_i$ corresponding to $\#i$, and mapped to its latent embedding using the image encoder. Remaining text elements of $goal$ are embedded using the text encoder. The embeddings are concatenated and processed using a Transformer followed by an MLP to obtain the state-sentence agreement value. When a goal text contains multiple sentences, we process each sentence separately, and take the minimum over the resulting agreement values (high agreement with the full text should imply high agreement with all constituent sentences).

To produce training data for the GMM, we generate sentences from the object configuration data explained in section 4A. Table 1 lists the sentence patterns generated, the number of examples obtained for each pattern, and the meaning of each sentence type. As is evident from Table 1, the number of examples differs substantially per sentence pattern (up to a factor 39). To avoid detrimental effects of this skew on accuracy, we oversample from the rarer pattern types during batch generation to equalise exposure to the sentence patterns.

In addition to true state-sentence pairs, training also requires false pairs. We generate false pairs in two ways. The first is to simply pair unrelated states and sentences. The second is to make small modifications to the sentence so as to render it untrue. For example, a sentence such as *"#1 is on #2."* may be rewritten into *"#2 is on #1."* to construct a false pair with the same elements as the original true pair. This promotes acquisition of the connections between grammatical and spatial relations. Finally, to robustify against imperfections in predicted state images, we apply blur and noise augmentations to the state images.

### *E. Image Conversion Module (ICM)*

The modules discussed above are sufficient for planning in the simulated manipulation scene. However, for real-world operation we need one additional module. Recall that the modules comprising the planning system are trained on simulation data. The use of simulation data allows us to easily vary object layouts and shape parametrisations. However, training on simulation data introduces a need to bridge the reality gap when we wish to perform real-world robotic manipulation using the trained system. For this purpose, we introduce domain adaptation by means of image conversion. This is the role of the Image Conversion Module (ICM).

Fig. 5 illustrates the concept. Core of the module is a neural network that takes an RGBD image from the real-world setup as input, and outputs an RGBD image of the same scene represented in the visual style of the simulation environment. We can then use this *sim-style* image as input to the sim-trained planning system. This circumvents the need to collect large amounts of real-world data for training the planning system. The network architecture used for the ICM is a Swin-Unet [22] modified to process 4-channel image data. Before conversion to sim-style, real world images are converted into orthogonal projection, in order to match the projection of simulation images and eliminate dependence of objects' appearance on the camera's position in the XY plane. The depth channel of the real-world image locates each pixel in 3D space, allowing us to remap the pixels into orthographic projection. Remapping will leave some pixels in the resulting image without content, in particular pixels on the workspace surface that are occluded by scene objects in perspective projection but not in orthographic projection. To fill in these pixels we prepare a background RGBD image of the real-world setup *without* any scene objects, and convert it into orthogonal projection. Some scene objects can appear incomplete in the resulting image, but we find that the conversion network is capable of inpainting the missing parts during image conversion.

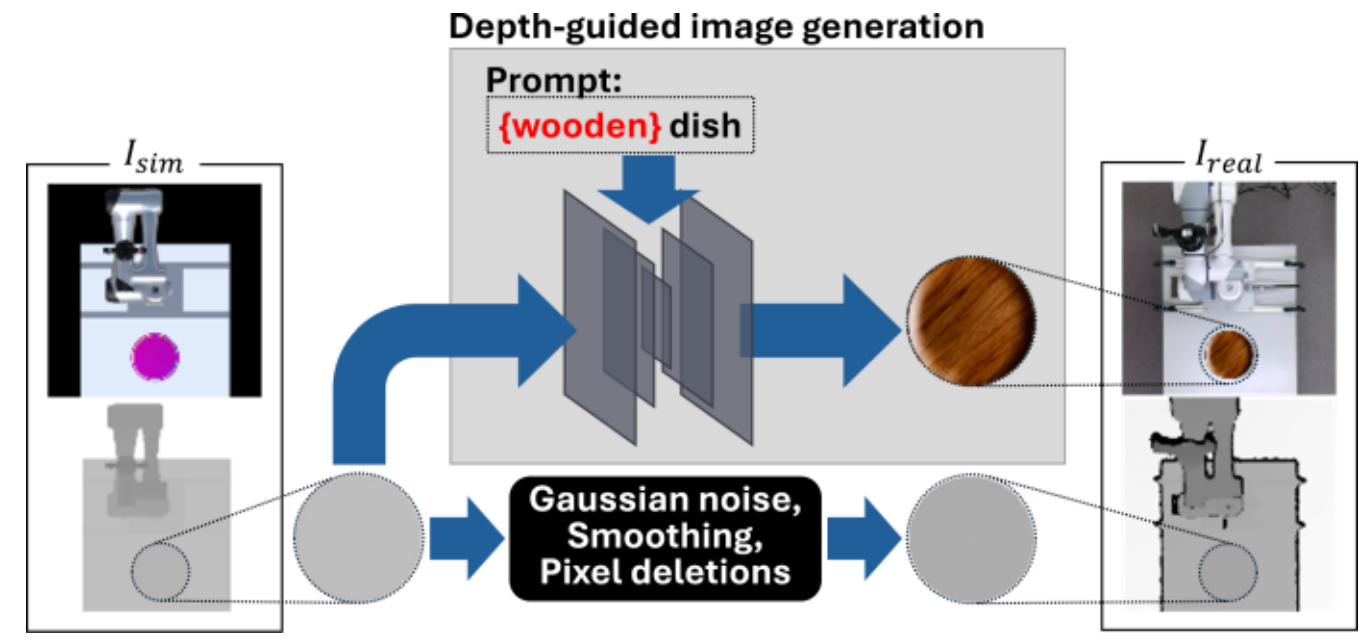

Fig. 6. Pipeline for generating ICM training data.

Next we discuss data generation for training the ICM. We require pairs ($I_{real} \in \mathbb{R}^{H \times W \times 4}, I_{sim} \in \mathbb{R}^{H \times W \times 4}$) of *real-style* and *sim-style* images. The images marked *Real RGBD* and *Sim RGBD* in Fig. 5 provide an example. We denote the RGB and D parts comprising these images as follows:

$$I_{real} = [I_{real}^{RGB}, I_{real}^{D}], I_{sim} = [I_{sim}^{RGB}, I_{sim}^{D}].$$

The images in a pair should depict matching objects in matching poses. In other words, the images should differ in visual style only, while depicting the same scene content. Furthermore, to enable the planning system to generalise over variation in objects' visual appearance, it is desirable that the *real-style* images ($I_{real}$) present objects in varied colours and materials. However, collecting a large and varied set of physical objects for generating this data is costly and impractical. Here we use a pre-trained generative model to synthesise visually varied objects.

Various methods for synthetic training image data have been proposed, such as *GenAug* [23] and *Grasp Anything* [24]. However, our use case imposes the constraint that the shape of the generated objects should closely match the shapes of the objects in $I_{sim}$. To satisfy this constraint, we adopt the combination of Stable Diffusion [25] with ControlNet [26] to perform object image generation conditioned on depth images.

The pipeline for generating image pair data is illustrated in

Fig. 6. Associated with each image $I_{sim}$ in the simulation dataset is a set of isolated depth images of all objects depicted in $I_{sim}$. We map pairs of object depth image and text prompts to realistic synthetic RGB images using ControlNet and Stable Diffusion. The depth image constrains the generated object to be consistent in shape with the object in simulation, and inclusion of the object class name (*dish, cup, block, ball*) promotes visual consistency with the object type. Variation in object appearance is obtained by including various colour and material descriptors (e.g. *red, ceramic*) in the text prompts. These descriptors are drawn randomly from a prepared list of 30 words. By scaling and pasting the resulting objects onto the aforementioned background image at the same coordinates where the corresponding objects are located in $I_{sim}$ we can synthesise a realistic $I_{real}^{RGB}$ with content matching $I_{sim}$.

Synthesising $I_{real}^{D}$ is comparatively straightforward, as objects' appearance in the depth channels is less variable. We synthesise $I_{real}^{D}$ by scaling and pasting the object depth images into the depth channel of the background image. To improve robustness against noise and missing pixel data in real-world depth data, we apply Gaussian noise, smoothing, and random pixel deletions. The above procedure lets us generate $(I_{real}, I_{sim})$ pairs in a fully automated manner.

To weed out low-quality examples, we add an additional quality control step. We found that occasionally, images are generated that do not match the description given in the prompt. Such examples may negatively affect ICM training. To eliminate such examples, we filter the generated data using a One-Class SVM [27], an unsupervised method for identifying outliers. For each prompt used during data generation, we apply the One-Class SVM to the set of images generated using that prompt, and discard the examples that are classified as outliers. The proportion of outliers (One-Class SVM parameter ν) is set to 0.03. For purpose of evaluation, we labelled the 330 examples generated for the prompt "blue dish" by visual inspection. Filtering reduced the proportion of invalid examples from 2.7% to 0.9%, discarding 6/9 invalid examples and 1/321 valid examples.

The composition of the filtered object dataset is shown in Table 2. Using this object data, we generate 100k $(I_{real}, I_{sim})$ pairs as training data and another 10k pairs as validation data. We then train the ICM on the resulting dataset. Data augmentations used in ICM training are image translation and rotation, colour shifts, Gaussian noise, partial deletions, and addition of artificial shadows using the Albumentations [28] library. Furthermore, we rescale depth values to emphasise depth contrast in the range where scene objects reside.

## MODULE EVALUATIONS

Before moving on to the planning logic and evaluation of the integrated planning system, we evaluate the trained system modules in isolation.

### *A. Affordance Recognition Module Evaluation*

We let the ARM process all states in the test and training datasets, and check how many of the ground truth affordances are recovered. We use a confidence threshold of 0.5, and tolerances of 2.5cm for coordinates and 5° for angles when matching recognised affordances to ground truth affordances. Duplicate affordances in ARM output are filtered. Affordances that do not match any of the ground truth affordances are counted as *spurious* recognitions. Table 3 reports the recogni-

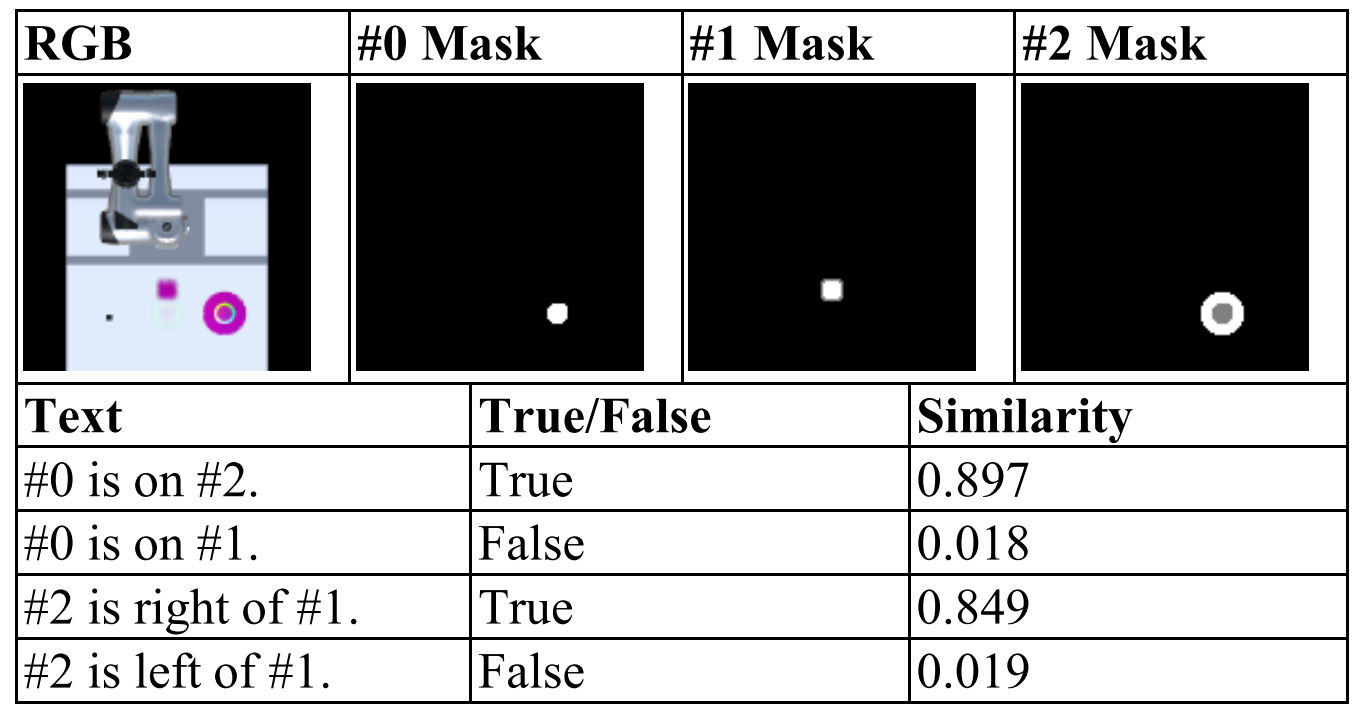

| Text | True/False | Similarity |
|---|---|---|
| #0 is on #2. | True | 0.897 |
| #0 is on #1. | False | 0.018 |
| #2 is right of #1. | True | 0.849 |
| #2 is left of #1. | False | 0.019 |

Fig. 7. Example of state-text agreement quantification by the goal matching module. Note the half-intensity area in the #2 mask indicating where the object is present but occluded.

Table 2. Composition of object dataset for ICM training.

| Object Type (view) | Example count |
|---|---|
| Cup (Top view) | 6208 |
| Cup (Side view) | 6024 |
| Cup (Bottom view) | 7700 |
| Plate (Top view) | 6632 |
| Ball | 6122 |
| Block | 6508 |

Table 3. Affordance recognition accuracy.

| | Grasp | Place | Push | Insert | All | Spurious |
|---|---|---|---|---|---|---|
| **Test** | 0.92 | 0.90 | 0.86 | 0.91 | 0.90 | 2.8/state |
| **Train** | 0.93 | 0.94 | 0.87 | 0.89 | 0.91 | 2.6/state |

Table 4. Prediction accuracy.

| Set | Step | All-area accuracy | | Changed area accuracy | |
|---|---|---|---|---|---|
| | | Mean (std) | Median | Mean (std) | Median |
| Test | 1 | .0031 (.0022) | .0033 | .042 (.044) | .031 |
| | 2 | .0052 (.0050) | .0051 | .049 (.046) | .041 |
| | 3 | .0070 (.0054) | .0064 | .058 (.043) | .046 |
| | 4 | .0090 (.0081) | .0077 | .065 (.047) | .056 |
| | 5 | .010 (.0075) | .0091 | .071 (.047) | .061 |
| | All | .0069 (.0065) | .0059 | .057 (.047) | .045 |
| Train | 1 | .0032 (.0028) | .0033 | .043 (.040) | .031 |
| | 2 | .0052 (.0044) | .0051 | .049 (.042) | .041 |
| | 3 | .0072 (.0057) | .0066 | .059 (.044) | .048 |
| | 4 | .0088 (.0064) | .0080 | .067 (.047) | .058 |
| | 5 | .010 (.0066) | .0091 | .070 (.046) | .059 |
| | All | .0069 (.0059) | .0061 | .058 (.045) | .046 |

Table 5. Goal matching accuracy.

| Sentence type | Acc. | Rec. | Prec. |
|---|---|---|---|
| {**#m**} is grasped. | 1.00 | 1.00 | 1.00 |
| {**#m**} is on {**#n**}. | 0.94 | 0.94 | 0.89 |
| {**#m**} is on the {row} {**col**}. | 0.99 | 0.98 | 0.99 |
| {**#m**} is {direction} {**#n**}. | 0.99 | 0.99 | 0.99 |
| Total | 0.98 | 0.98 | 0.98 |

Table 6. Image conversion accuracy on synthetic data.

| | SSIM | LPIPS | L1 (all) | L1 (objects) |
|---|---|---|---|---|
| **Test** | 0.9795 | 0.0153 | 1.0784 | 10.208 |
| **Train** | 0.9817 | 0.0026 | 0.9972 | 9.398 |

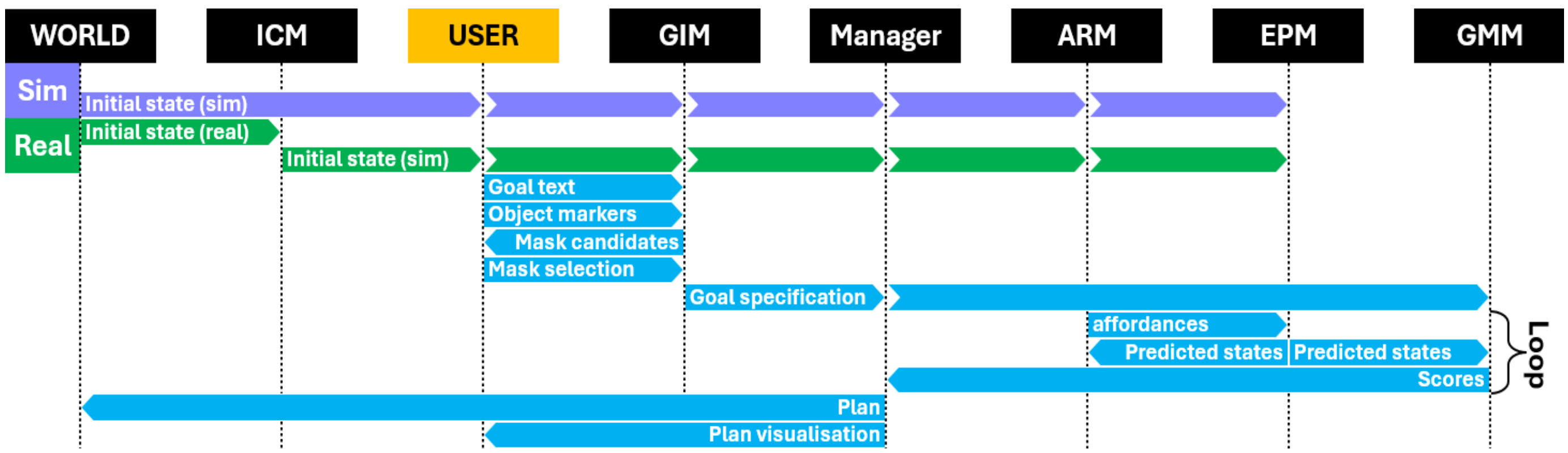

Fig. 8. Information flow during planning (top to bottom). ICM: Image Conversion Module. GIM: Goal Interpretation Module. ARM: Affordance Recognition Module. EPM: Effect Prediction Module. GMM: Goal Matching Module.

tion rate per affordances type and the mean number of spurious recognitions per state image. Unsurprisingly, the majority of recognition failures occur on edge cases, i.e. where the gripper can or cannot reach an object by a small margin. During data generation, we use a mock gripper of somewhat inflated size to check for collisions with some additional leeway, so false positives on edge cases are often in fact executable.

### *B. Effect Prediction Module Evaluation*

We evaluate the EPM in terms of the pixel-level error between predicted state images and ground truths (Table 4). Note that any single action leaves most pixels unchanged. Following [2], we also measure prediction error exclusively over the pixels that change from one state to the next. We observe mean pixel value errors of 0.0069 overall and 0.057 for changed pixels specifically.

### *C. Goal-Matching Module Evaluation*

Table 5 reports accuracy, recall and precision at a true/false threshold of 0.5, over a test data set of size 25764, split evenly between positive and negative examples. Negative examples are generated in the same manner as during training. Fig. 7 shows text-state agreement for various sentences on an example state. Overall accuracy is 98%. Most misclassifications occur on borderline cases, e.g. failure to recognise objects as stacked when their area of overlap is very small.

### *D. Image Conversion Module Evaluation*

We evaluate image conversion accuracy using LPIPS (Learned Perceptual Image Patch Similarity, [29]) and SSIM (Structural Similarity Index Measure, [30]) as our quality measures. Quantitative results on our synthetic dataset are reported in Table 6. We observe generally good conversion quality on synthetic data, although thin object edges can be challenging. Conversion results on images from the real-world setup can be found in the hardware experiments section (7B).

## VI. Planning Logic

Combining the above modules make it possible to perform manipulation planning on basis of text-based goals. Fig. 8 illustrates the process. First an image of the initial world state is obtained. For simulation experiments this is a *sim-style* RGBD image in orthogonal projection. For real-world experiments this is a *real-style* RGBD image in perspective projection. In the latter case, we apply perspective conversion followed by style conversion using the ICM. Then we process the state image and goal text using the GIM to obtain masks for the named objects and the standardised goal text.

The goal specification and initial RGBDMMM state are passed to a planning *manager* encapsulating the actual planning process. The initial state is passed on to the ARM and EPM. The former finds the list of available affordances in the current state, and passes the list to the EPM. Using the initial state and the affordance list, the EPM predicts the future state for each affordance.

These predictions are scored for agreement with the goal specification by the GMM. Recall that in mask channels, pixel values of 0.0, 0.5 and 1.0 represent object absence, occluded presence, and unoccluded presence. When forwarding predicted states to the GMM, we round mask channel pixel values to multiples of 0.5, thereby eliminating blur around object edges and low-intensity artifacts. Predictions are also passed on to the ARM to find the affordances in those future states. This process of recognition, prediction, and scoring is repeated up to a user-specified search depth. Finally, we return the sequence of affordances that produced the state prediction with the best goal agreement as our manipulation plan, along with a visualisation of this plan. This plan may be shorter than the user-specified search depth.

To reduce computational cost, we constrain the planning process to meaningfully distinct branches of the search tree, by applying a simple pruning process on the child nodes (predicted states) that are expanded from each parent node. Different affordances may produce qualitatively equivalent states. For example, grasping a cube at a 90° angle or a 270° angle will in many cases produce indistinguishable outcomes. Hence, for the set of child nodes expanded from each parent node, we measure the difference between the state predictions. If two predictions have fewer than 100 pixels differing by more than 0.15 on the RGB channels, we consider the states identical and discard the one whose affordance was recognised with lower confidence.

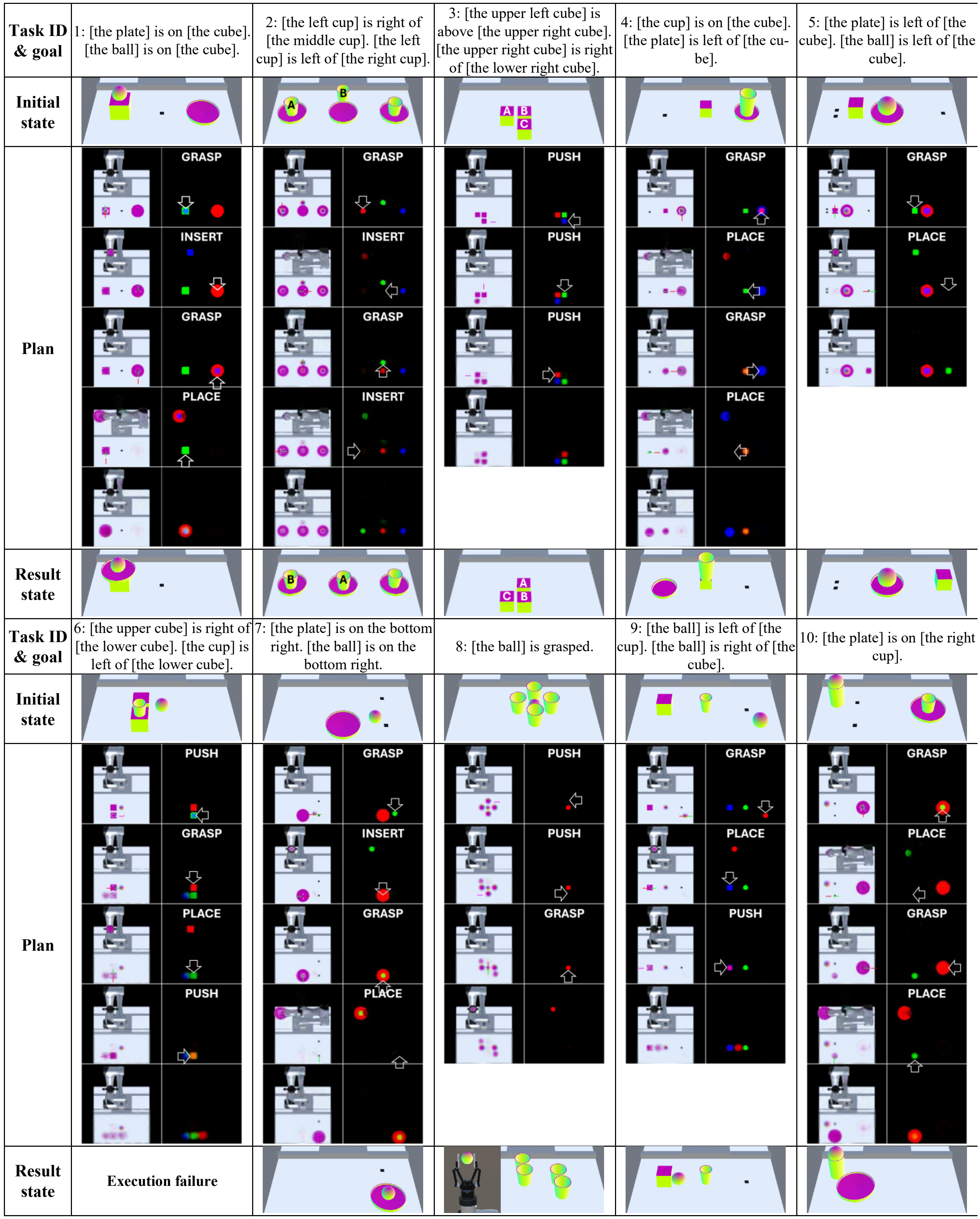

Fig. 9. Planning results for simulation experiments. In each plan, states after the first are system-generated predictions. Depth channel omitted. Right-side column of each plan shows object mask channels (colourised to RGB to improve distinguishability). See text for details.

Table 7. System configuration.

| CPU | AMD EPYC7702P (64 Core / 2.0 GHz) |
|---|---|
| GPU | NVIDIA H100 80GB |
| RAM | 512GB |
| ROS | ROS1 Noetic |
| Unity | 2021.3.38f1 |
| RGBD Camera | Azure Kinect |
| Arm | Universal Robots UR5 (simulation)<br>DOBOT CRA5 (hardware) |
| Gripper | Robotiq 2F-140 |

## VII. Planning Experiments

We evaluate the integrated planning system on a set of planning tasks designed to test the system's ability to handle various challenges. Note that the planning process is fully deterministic (repeat runs with the same goal and initial state produce identical plans). We let the virtual UR5 execute the generated plans, using MoveIt [31] to generate joint trajectories for the gripper coordinates specified in the plan. Table 7 provides a specification of the system configuration (hardware and software) used for plan generation.

### *A. Simulation Experiments*

Fig. 9 shows initial states, goals, generated plans, and obtained result states for our set of 10 simulation tasks. In each plan, the state images beyond the first are predictions generated by the EPM. Mask channels corresponding to objects #1, #2, and #3 are coloured in red, green and blue for ease of distinction. The depth channel is omitted due to space constraints. Explanatory markers and text (white) are edited in manually. Letters in initial and result states are added to distinguish visually identical objects, and are not visible to the planning system. Recall that small black markers on the table surface indicate placement positions. We briefly explain the thinking behind the task designs below.

Task 1: The goal here is to position both the ball and the plate on the cube. The stacking order for the ball and plate is not specified, and the ball is already on the cube. A naïve planner might fail by trying to place the plate on the ball. The correct solution is to move the ball onto the plate, and then move the plate-ball compound onto the cube. This requires temporarily undoing a goal condition (ball-on-cube) that holds in the initial state, exploitation of the effect that an object on a plate moves along with that plate, and retention of the position of the cube as it becomes occluded by the plate (to match the solution state to the goal).

Task 2: For a lack of placement positions (black markers) on the table surface, cups can only be placed on plates here. To satisfy the goal text, cup A must be placed on the middle plate, and cup B must be placed on the left plate. However, the left plate is initially occupied by cup A, so the choice of cup to grasp first must be informed by foresight of the subsequently available PLACE affordances.

Task 3: This block-pushing puzzle admits exactly one valid order of push actions.

Task 4: The goal specifies target positions for the cup and plate relative to the block. The cup (which is large relative to the block) must be on top of the block, which implies that the block ends up fully occluded when the goal is satisfied. Hence, recognising that the goal is satisfied requires the system to keep track of the whereabouts of a fully occluded object, and assess positions of other objects relative to the occluded object.

Task 5: The goal requires both the ball and the plate to be to the left of the block, and there are two placement positions to the left of the block. However, the sizes of the objects and the close proximity of the placement positions to the block make it impossible to place the plate to the left of the block without colliding with it. The ball, however, could be placed. The solution is to satisfy the goal not by moving the ball and the plate but by instead moving the block to the right of the other objects.

Task 6: A specific alignment of objects must be obtained in absence of placement positions on the table surface. Our intended solution here was to push the lower cube to the left, then push the cup off to the left, then push the upper cube down. However, the system invented a different, unintended solution, exploiting the fact that the upper cube can be brought to the right of the lower cube by placing it on the upper cube and then pushing it off to the right. This strategy may have worked, but the initial push affordance is positioned too low in the Z dimension, leading to execution failure. Note that the state predictions for this task are quite fuzzy. This is likely due to the fact that the effect of pushing an object off the top of another object is somewhat stochastic.

Task 7: This task requires both the ball and the plate to be positioned on the lower right of the table. The ball already satisfies this requirement in the initial state, but also obstructs the plate from being pushed to or placed in the target area. Similar to task 1, the solution requires temporary undoing of a goal condition and exploitation of the effect that an object on a plate moves along with that plate.

Task 8: This task is cleared by grasping the ball, but the gripper's access to the ball is obstructed by the surrounding cups. Solving this task requires consideration of how affordance executions affect availability of future affordances: to reveal a GRASP affordance on the ball, two cups *on opposing sides* of the cup must be pushed aside, such that both fingers of the gripper can reach the ball.

Task 9: Solving this task requires exploitation of the fact that objects can be brought side-by-side by placing one on top of another and then pushing the top one off. The choice of the bottom object is important here: both the block and the cup are in viable positions, but placing the ball in the cup would make it impossible to push the ball off.

Task 10: The plate should be brought on top of the cup standing on top of it. This requires moving the cup on the table surface (onto one of the placement markers), and setting the plate on top of it. The additional catch here is that there is a second cup standing near one of the placement positions, that is taller than the right cup. The right cup could be placed there, but the taller cup would prevent placement of the plate on top of the right cup there. Hence the correct choice of placement position requires some foresight. Additionally, the right cup will necessarily be fully occluded in any state satisfying the goal, so in this task too, tracking the whereabouts of an occluded object is essential.

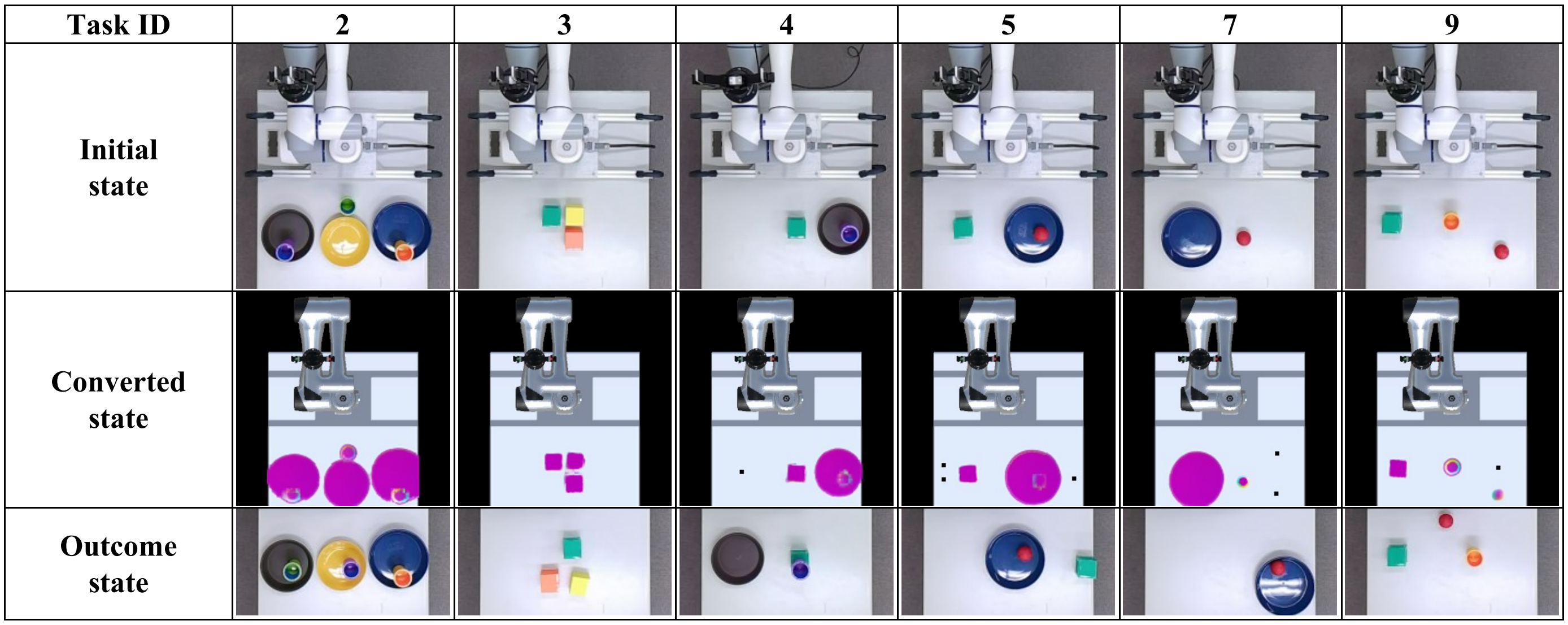

Fig. 10. Results of hardware experiments. Depth channel omitted. The *second* row shows initial states converted to sim-style using the ICM.

We see that the system generates sensible plans for 9 out of our 10 tasks, successfully avoiding our various traps.

### *B. Real-world Experiments*

Next, we perform a similar set of experiments using our real-world recreation of the experimental setup (Fig. 1, right panel). In this setup, the ICM comes into play, converting real-world images into sim-style images and enabling generalisation over object colours and textures. As the real-world context increases stochastic elements, we run 5 trials per task. Table 8 reports success rates per task. Half of the tasks are solved consistently. For tasks with at least one successful trial, Fig. 10 shows the initial, converted, and outcome states for one trial. Note the varied appearance of the scene objects. The ICM maps visually varied objects to a consistent appearance matching the simulation environment.

Failure patterns are as follows. The ICM struggles with some cases where objects are stacked in the initial real-world state, leading to consistent failure on tasks 1 and 6. In task 3 and 10, we varied the position of the cup on the plate. We see prediction failures for cup grasping actions when the cup is positioned out-of-centre on a plate in the initial state. Our data generation system produces configurations with stacked objects, but the top object in such configurations is almost always centred on the bottom object. Hence, out-of-centre states are unfamiliar territory. In task 8, the ARM erroneously recognises the ball as graspable after moving just one cup out of the way.

## VIII. Discussion & Conclusion

Our experiments show that the system is capable of solving tasks requiring multi-step foresight, “detouring” (temporarily undoing already-satisfied goal conditions), consideration of secondary affordance effects (e.g. if object A is on B, A moves with B), and extensive occlusion. Domain adaptation lets us transfer these abilities to real-world objects of varied appearance. However, we did observe a notable drop in success rates on real-world experiments.

The full system combines a number of different modules, which were trained separately in the present work. As discussed in Section 5 and the appendix, we apply various forms of data augmentation to improve each module’s robustness to imperfections in the data produced by the module feeding it, but it is difficult to fully cover the characteristic imperfections of each module. The EPM, for example, is trained on simulation data, but in hardware experiments it has to process images produced by the ICM, which contain noise and artifacts not fully covered by the various augmentations applied during EPM training. Reduced prediction accuracy on converted images was a major contributor to failures on real-world tasks. We expect that improvements could be obtained by training each module on data that has been passed through its feeding module. We did not pursue this strategy here, however, as it would incur extensive additional data processing and impose a fixed order on the training processes of the various modules.

Another open issue is planning time cost. In [2], planning time costs were on the order of a few seconds, but the present system is significantly slower, taking between tens of seconds to a few minutes per task. The introduction of a multimodal NN for goal matching (replacing the sliding window image region matching employed in [2]) adds substantial computational cost, and the use of a dynamic viewpoint in prediction incurs some additional computational costs as well. Another speed-limiting factor in our present implementation is the mixing of frameworks. Most modules are built on existing code bases, which were in turn built on different frameworks. The ARM, GMM, and ICM are implemented in PyTorch [32], whereas the EPM and additional processing in the planning process are implemented in JAX [33]. This results in frequent data format conversions and data transfers between CPU and GPU that could be avoided in a single-framework implementation.

Table 8. Success rates per task in hardware experiments.

| Task | 1 | 2 | 3 | 4 | 5 | 6 | 7 | 8 | 9 | 10 |
|---|---|---|---|---|---|---|---|---|---|---|
| Success rate | 0/5 | 5/5 | 5/5 | 1/5 | 5/5 | 0/5 | 5/5 | 0/5 | 5/5 | 0/5 |

As noted in the introduction, and closely related to the issue of time efficiency, human planning combines a sub-symbolic understanding of real-world dynamics with symbolic reasoning. We focused here on the former, but hybridisation with the latter could greatly improve planning efficiency. Symbolic reasoning facilitates efficient search for candidate plans, while sub-symbolic reasoning allows for verification of candidate plans at fine granularity, and for broad “outside the box” plan search when symbolic search fail to produce a viable plan (cf. “non-canonical affordance effects” in [2]). Furthermore, it may be possible to acquire symbolic rules by discovering consistent patterns in learned sub-symbolic dynamics. Exploring the potential of fine-grained prediction as a substrate for the grounding and discovery of symbolic rules remains as future work.

## Appendix

Here we specify the architectures and hyperparameters of the neural network architectures used in the various modules of the planning system.

### A. Affordance Recognition Module (ARM)

The affordance recognition network is a modification of ScaledYOLOv4 [19] with the architecture given in Table 9. The architecture was adjusted to produce a finer grid of candidate recognitions than the original settings, because we may need to recognise multiple exemplars of the same affordance type in close proximity.

Table 9. Affordance Recognition Network.

| | # | Layer type | Channels, kernel, stride, upscale | Input layer(s) |
|---|---|---|---|---|
| backbone | 1 | conv. | 3→16, 3, 1, - | |
| | 2 | conv., 1xCSP | 32, 3, 2, - | 1 |
| | 3 | conv., 2xCSP | 64, 3, 2, - | 2 |
| | 4 | conv., 8xCSP | 128, 3, 2, - | 3 |
| | 5 | conv., 8xCSP | 256, 3, 2, - | 4 |
| | 6 | conv., 4xCSP | 512, 3, 2, - | 5 |
| Head | 7 | CSPSPP | 256 | 6 |
| | 8a | conv. | 128, 1, -, - | 5 |
| | 8b | conv. | 128, 1, -, - | 7 |
| | 9 | concat., 2x rCSP | 128 | 8a, 8b |
| | 10a | conv. | 64, 1, -, 2 | 4 |
| | 10b | conv. | 64, 1, -, 4 | 9 |
| | 11 | concat., 2x rCSP | 128 | 10a, 10b |
| | 12 | conv. | 128, 3, -, - | 11 |
| | 13 | detection | - | 12 |

We replace the bounding box width and height outputs with a single affordance angle output, using 8 equally spaced anchor values. Angles are normalised to the $[0, 1)$ range. We define the angle loss as follows:

$$loss_{angle} = min(d, 1 - d)$$
$$d = (\hat{a} - a) \bmod 1$$

where $a$ is the ground truth angle and $\hat{a}$ the value given by the network. Other loss definitions follow ScaledYOLOv4. ARM training employs the following data augmentations:

1. Shift augmentation: we apply random shifts of -8 to +8 pixels along the X and Y axes, with corresponding adjustments in affordance coordinates.
2. Noise augmentation: we apply uniform noise from range $[-r, r]$, where $r$ is drawn randomly per example from $[0,10]$, and clip the result to $[0,255]$.
3. Artifact augmentation: we blend the state image with the previous state image from the same data sequence at a random intensity drawn per pixel from $[0, 0.25)$.

### B. Effect Prediction Module (EPM)

Table 10 specifies the architecture for the EPM, which is based on the prediction network of our previous work [2]. One pass through this architecture corresponds to prediction for one affordance. Separate instances of the $P$ submodule are used for different affordance types.

Table 10. Effect Prediction Network Hyperparameters.

| | Layer # | Details | Activation |
|---|---|---|---|
| **E** | 1 conv. | Channels: $14 \to 8$<br>Kernel: $3 \times 3$<br>Stride: $2 \times 2$ | $tanh(x)$ |
| | 2 conv. | Channels: $8 \to 8$<br>Kernel: $3 \times 3$<br>Stride: $2 \times 2$ | |
| | 3 linear | $8192 \to 8192$ | |
| | 4 linear | $8192 \to 4096$ | |
| | 5 linear | $4096 \to 256$ | |
| | - | Local response normalisation | |
| **P** | 1~8 | $4 \times$ Mixed connectivity block (B in Fig. 3) | $tanh(x)$ |
| | - | Local response normalisation | |
| **D** | 1 linear | $8192 \to 8192$ | $tanh(x)$ |
| | 2 linear | $8192 \to 4096$ | |
| | 3 linear | $4096 \to 256$ | |
| | 4 conv. | Upscale: $2 \times 2$<br>Channels: $8 \to (14 + 2)$<br>Kernel: $3 \times 3$ | |
| | 5 conv. | Upscale: $2 \times 2$<br>Channels: $8 \to 8$<br>Kernel: $3 \times 3$ | $clip(x, 0, 1)$ |

*Local response normalisation* refers to the following simplified normalisation:

$$LRN(x) = \frac{x}{\sqrt{\|x^2\|_1}}$$

The training loss consists of separate components for RGBD channels and mask channels. The RGBD loss $loss^{RGBD}$ is simply the mean squared error over prediction and ground truth, averaged over all states in the sequence. Mask channels are dominated by zero values, making mask prediction prone to falling into the local minimum of prediction zeroes exclusively. To focus the loss on *changes* in the mask channels, we multiply per-pixel loss for pixels that have changed relative to the preceding state by $w^{changed}$, and loss for pixels that stayed the same by by $w^{same}$, where:

$$w^{changed} = \frac{b}{d}, \qquad w^{same} = \frac{1-b}{1-d},$$

with $b = 0.1$ and $d$ the proportion of changed mask-channel pixels between the state and its preceding state.

We train the network using the signSGD [34] loss, with an adaptively decreasing learning rate initialised to $5 \times 10^{-5}$. We evaluate accuracy on the validation set every 10k iterations, and when accuracy fails to improve for 5 evaluations in a row, the learning rate is reduced by a factor 2. Training is terminated when the learning rate falls below $10^{-10}$.

EPM training employs the following data augmentations:

1. Mirroring of all states and affordances in a sequence.
2. Affordance position & angle noise
3. Rotating affordance angles by 180° for affordances where this produces identical outcomes.
4. Blacking out random patches of the input state areas that remain constant over the state sequence.

*C. Goal Matching Module (GMM)*

The GMM reuses parts of the CLIP [21] architecture, albeit in a different configuration and scaled down somewhat. Table 11 specifies the architecture.

Table 11. Goal Matching Network Hyperparameters.

| **Image encoder** | | Activation |
|---|---|---|
| 1x 2D convolution | Kernel: 32x32<br>Input channels: 1<br>Output channels: 768 | $GELU$ [35] |
| 8x Residual Attention | Heads: 16<br>Output dim: 512 | |
| 2x Linear | Dims: 512→512 | |
| **Transformer** | | |
| 8x Residual Attention | Heads: 16<br>Output dim: 512 | $GELU$ |
| **MLP** | | |
| 3x Linear | Dims: 512→256→128→1 | $GELU$ |

The goal matching network is trained using the following loss:

$$loss = \frac{1}{n_{batch}} \sum_{i=1}^{i=n_{batch}} \alpha \cdot (1 - e^{l_i})^{\gamma} \cdot l_i$$

$$l_i = y_i \cdot \log \sigma(x_i) + (1 + y_i) \cdot \log\big(1 - \sigma(x_i)\big)$$

where $x_i$ and $y_i$ are the output and ground truth for the $i^{th}$ example in the batch, batch size $n_{batch} = 16$, $\alpha = 1$, and $\gamma = 2$. The network is trained using the Adam optimiser with hyperparameters $\beta_1 = 0.9, \beta_2 = 0.99, \varepsilon = 10^{-7}$, and a weight decay of $10^{-7}$. We use a cosine learning rate schedule that starts at $10^{-7}$, gradually increases to $10^{-5}$ over the first epoch, and then gradually decays again. We train for 50 epochs in total.

*D. Image Conversion Module (ICM)*

The ICM adopts the Swin-Unet architecture [22] as-is, except for an increase in input and output channels from 3 to 4, in order to accommodate the depth channel of our RGBD imagery. The net is trained using the Adam optimiser with hyperparameters $\beta_1 = 0.9, \beta_2 = 0.999, \varepsilon = 10^{-8}$, weight decay of 0, and a fixed learning rate of $10^{-4}$. We train for 50 epochs with a batch size of 16.